\documentclass[journal,twoside,web]{ieeecolor}
\usepackage{tmi}
\usepackage{cite}
\usepackage{amsmath,amssymb,amsfonts}
\usepackage{algorithmic}
\usepackage{graphicx}
\usepackage{textcomp}
\usepackage{epstopdf}
\usepackage{multirow}

\usepackage{algorithm}
\usepackage{algorithmic}

\def\BibTeX{{\rm B\kern-.05em{\sc i\kern-.025em b}\kern-.08em
    T\kern-.1667em\lower.7ex\hbox{E}\kern-.125emX}}
\begin{document}
\title{A Teacher-Student Framework for Semi-supervised Medical Image Segmentation From Mixed Supervision}
\author{Liyan Sun, Jianxiong Wu, Xinghao Ding, Yue Huang, Guisheng Wang and Yizhou Yu, \IEEEmembership{Fellow, IEEE}
\thanks{The work is supported in part by National Key Research and Development Program of China (No. 2019YFC0118104), in part of ZheJiang Province Key Research Development Program (No. 2020C03073), in part by National Natural Science Foundation of China under Grants 81671766, 61971369, U19B2031, U1605252, 61671309, in part by Open Fund of Science and Technology on Automatic Target Recognition Laboratory 6142503190202, in part by Fundamental Research Funds for the Central Universities 20720180059, 20720190116, 20720200003, and in part by Tencent Open Fund.}
\thanks{Liyan Sun, Jianxiong Wu, Xinghao Ding, and Yue Huang are with the
School of Informatics, Xiamen University, Xiamen 361005, China (e-mail:
dxh@xmu.edu.cn). }
\thanks{Guisheng Wang is with the Department of Radiology, the Third Medical Centre, Chinese PLA General Hospital, Beijing, China
(e-mail: wanggs1996@tom.com)}
\thanks{Yizhou Yu are with the Deepwise AI Laboratory, Beijing 100125, China
(e-mail: yizhouy@acm.org)}
}

\maketitle

\begin{abstract}
  Standard segmentation of medical images based on full-supervised convolutional networks demands accurate dense annotations. Such learning framework is built on laborious manual annotation with restrict demands for expertise, leading to insufficient high-quality labels. To overcome such limitation and exploit massive weakly labeled data, we relaxed the rigid labeling requirement and developed a semi-supervised learning framework based on a teacher-student fashion for organ and lesion segmentation with partial dense-labeled supervision and supplementary loose bounding-box supervision which are easier to acquire. Observing the geometrical relation of an organ and its inner lesions in most cases, we propose a hierarchical organ-to-lesion (O2L) attention module in a teacher segmentor to produce pseudo-labels. Then a student segmentor is trained with combinations of manual-labeled and pseudo-labeled annotations. We further proposed a localization branch realized via an aggregation of high-level features in a deep decoder to predict locations of organ and lesion, which enriches student segmentor with precise localization information. We validated each design in our model on LiTS challenge datasets by ablation study and showed its state-of-the-art performance compared with recent methods. We show our model is robust to the quality of bounding box and achieves comparable performance compared with full-supervised learning methods.
\end{abstract}


\section{Introduction}
\label{sec:introduction}
\IEEEPARstart{A}{utomatic} and accurate segmentation of tissues, organs or lesions \cite{ref1,ref2,ref3,ref4,ref5,ref28} provides valuable biomarkers for diagnosis, treatment and prognosis. Deep convolutional neural networks (CNN) were proposed and proven to achieve efficient and accurate medical image segmentation. However, conventional CNN models are conditioned on massive paired medical images and their corresponding dense-labeled annotations, which imposes limitations on their utilities in practical clinical scenarios. To relax such exacting demands for large amounts of well-annotated densely label data pairs in medical image analysis, we consider a mixed-supervision condition where segmentation labels are only available for a small portion in medical image datasets while a large number of loose annotations in the form of bounding-box exist \cite{ref6,ref7,ref8,ref9} as shown in Figure. \ref{Fig1}. Please refer to Section. \ref{sec:PF} for detailed explanations of notations in this figure. A rationale for such assumption is that medical images with bound-boxing annotations are more easily obtained. Radiologists and trained experts are able to label cheap bounding box regions more efficiently than pixel-wise segmentation labels. However, such a weakly- and semi-supervised learning is still a challenging task due to three reasons:

\begin{figure}[t]
	\centering
	\includegraphics[width = 0.44\textwidth]{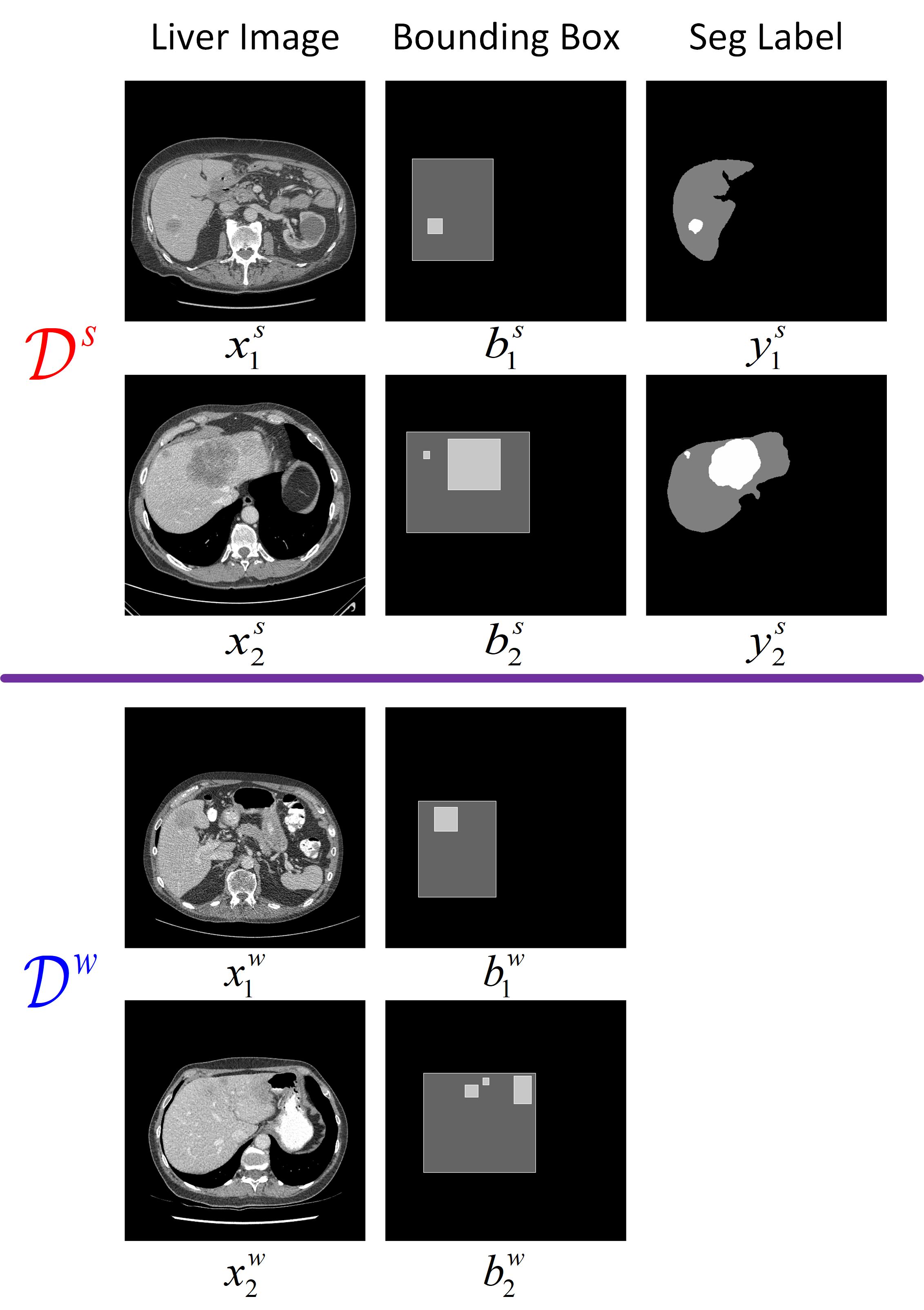}
	\caption{Two illustrated examples of data triplet in a dense-labeled dataset called ${\mathcal{D}^s}$ and data pair in a loosely-labeled dataset called ${\mathcal{D}^w}$. Here liver regions and interior lesions are shown on CT slices.}
	\label{Fig1}
\end{figure}

(1) \textit{How to leverage bounding-box labels more sufficiently.}. The supervision information in a segmentation label includes size, position and shape. Compared to a dense pixel-wise annotation, a typical rectangle bounding box has little shape information as illustrated in Figure. \ref{Fig1}, where bounding boxes merely enclose livers of various shapes in abdomen CT slices. Some previous methods coupled the network training from strong and weak supervision into one unified backbone \cite{ref6,ref7,ref8} or separate branches in a one-stage multi-task fashion \cite{ref9}. In training, each forward is depended on whether the input is strongly supervised or weakly supervised. However, directly utilizing ``crude" bounding-box labels without effective refinements poses limited improvements on segmentation. Guided by the limited number of dense supervision and bound box annotations, inferring shapes of an organ and its inner lesions improves the utility of the mixed-supervision dataset.

(2) \textit{How to develop weakly- and semi-supervised learning framework tailored for medical images.} In knowledge distillation \cite{ref27}, a trained teacher (also called annotator) model produces pseudo-labels on unlabeled or weakly-labeled data to train a student model. Although such a framework enables refinements on loose bounding-box labels in tasks of natural images \cite{ref10,ref11,ref12,ref13}, to our knowledge, few relevant studies are tailored to medical images. Compared to segmentations of general natural images, the relative spatial positions of segmented objects within one medical image is much less diverse. For example, a person may ride a horse on its back or stand by a horse feeding it. In contrast, as we shown in Figure. \ref{Fig1}, it is often the case that a lesion grows and lies inside an organ, forming a more regular and consistent geometrical pattern. Such idiosyncratic property in medical image segmentation calls for specific model designs for a teacher and student.

(3) \textit{How to better localize objects with mixed annotations}. In Figure \ref{Fig1}, we observe lesions within a liver are highly diverse in size. In many cases lesion regions are very small compared with a whole slice or even the organ containing them. One of the premises for accurate segmentation of a small object is accurate localization. Thus exploiting efficient architecture to facilitate accurate localization with cheap bounding-box labels is able to boost the ability of a segmentation model in small object segmentation.

To cope with these challenges, we proposed a teacher-student framework tailored for medical image segmentation from mixed supervision under a semi-supervised learning condition. The teacher-student framework enables shape refinement on a weakly annotated dataset. A hierarchical organ-to-lesion attention scheme is proposed to enhance a teacher annotator using bounding-box labels. Then a combination of pseudo labels produced by a teacher and partial manually-labeled annotations is used to train a student segmentor. In a localization branch, multi-scale features in a student segmentor are aggregated to predict positions of organ and lesion to improve its ability for localization without introducing extra parameters during testing.
Experiments on liver and lesion segmentation: LiTS dataset were conducted to demonstrate its state-of-the-art performance. We also demonstrate that our weakly- and semi-supervised learning framework approximates its full-supervised counterpart under certain circumstance and our model is shown to be robust to the bias in labeling bounding boxes.

\section{Related Works}
\label{sec:RelatedWorks}
\begin{figure}[t]
	\centering
	\includegraphics[width = 0.45\textwidth]{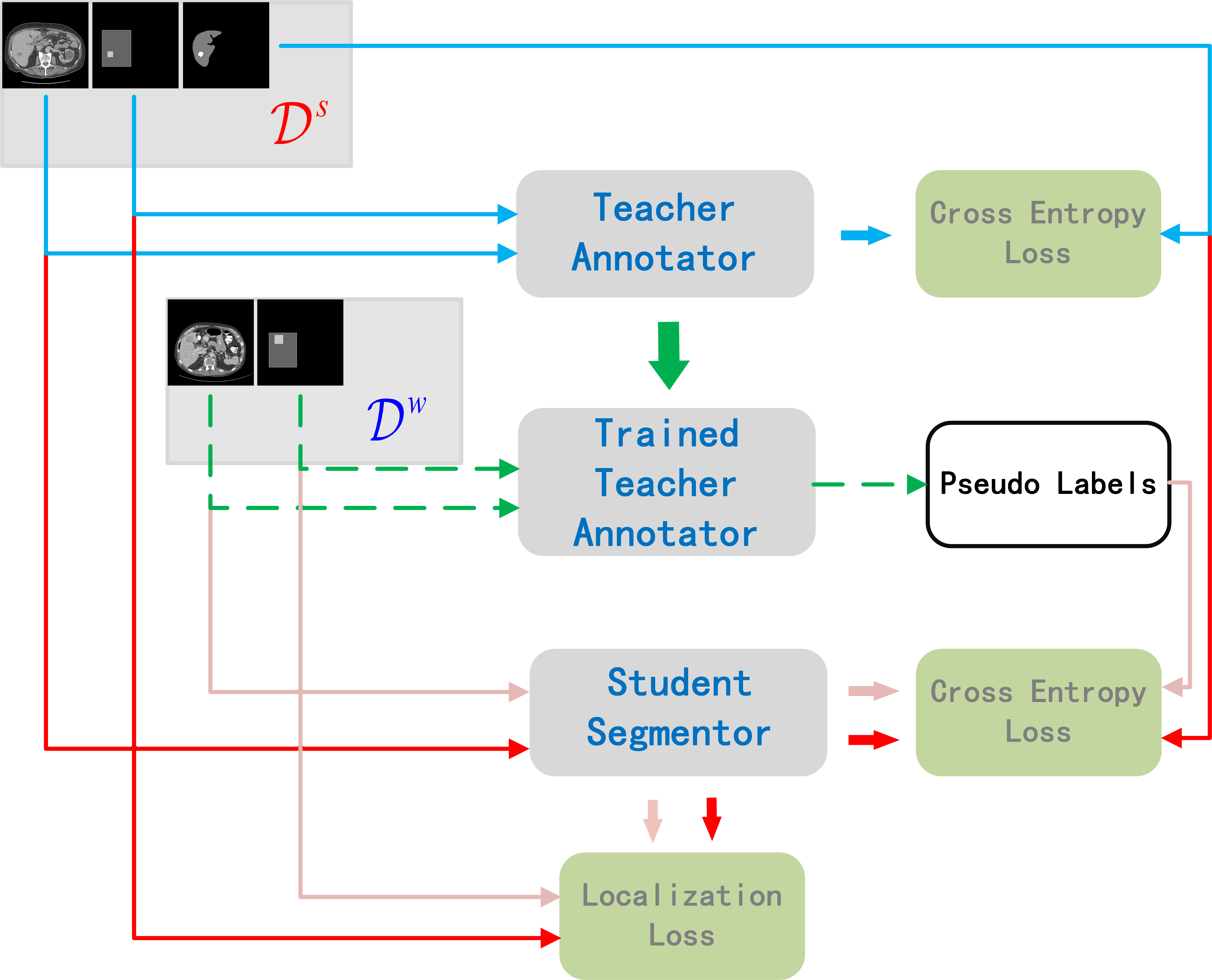}
	\caption{The workflow of the proposed semi-supervised learning. Blue lines represent training of a teacher annotator with a densely labeled dataset. Green dashed lines represent the automatic pseudo annotations produced by the trained teacher on a roughly labeled dataset. Then light and dark red lines together denote the training of student segmentor.}
	\label{Fig2}
\end{figure}
We first review the history of medical image segmentation based on full supervision. Then we discuss semi-supervised learning approaches relaxing demands for high-quality annotations in conventional methods.

\subsection{Medical Image Segmentation}
\begin{figure*}[t]
	\centering
	\includegraphics[width = 0.9\textwidth]{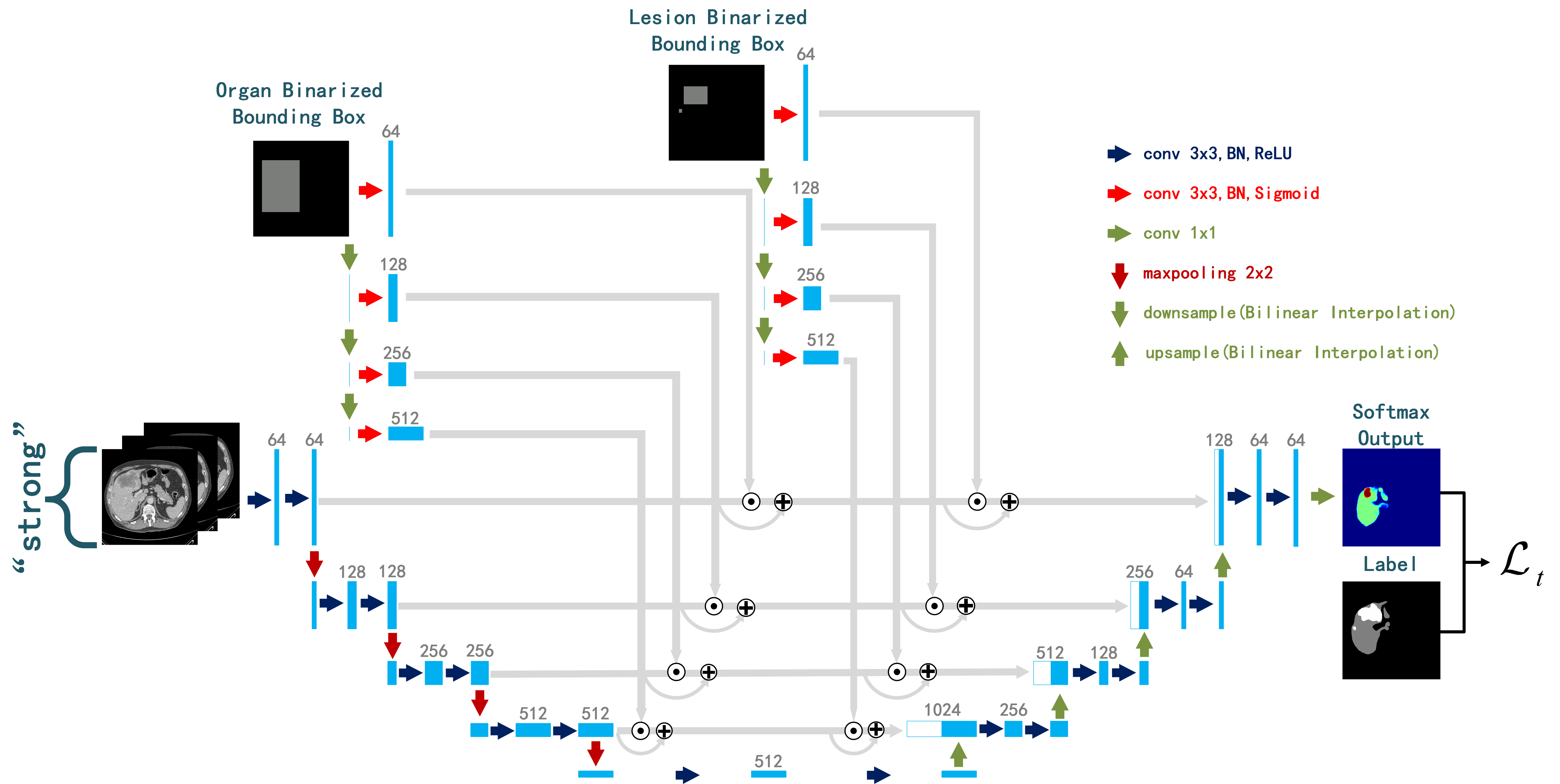}
	\caption{The deep network architecture of a teacher annotator with hierarchical organ-to-lesion attention weighting. Consecutive three slices are concatenated as network input.}
	\label{Fig3}
\end{figure*}

\begin{figure}[h]
	\centering
	\includegraphics[width = 0.4\textwidth]{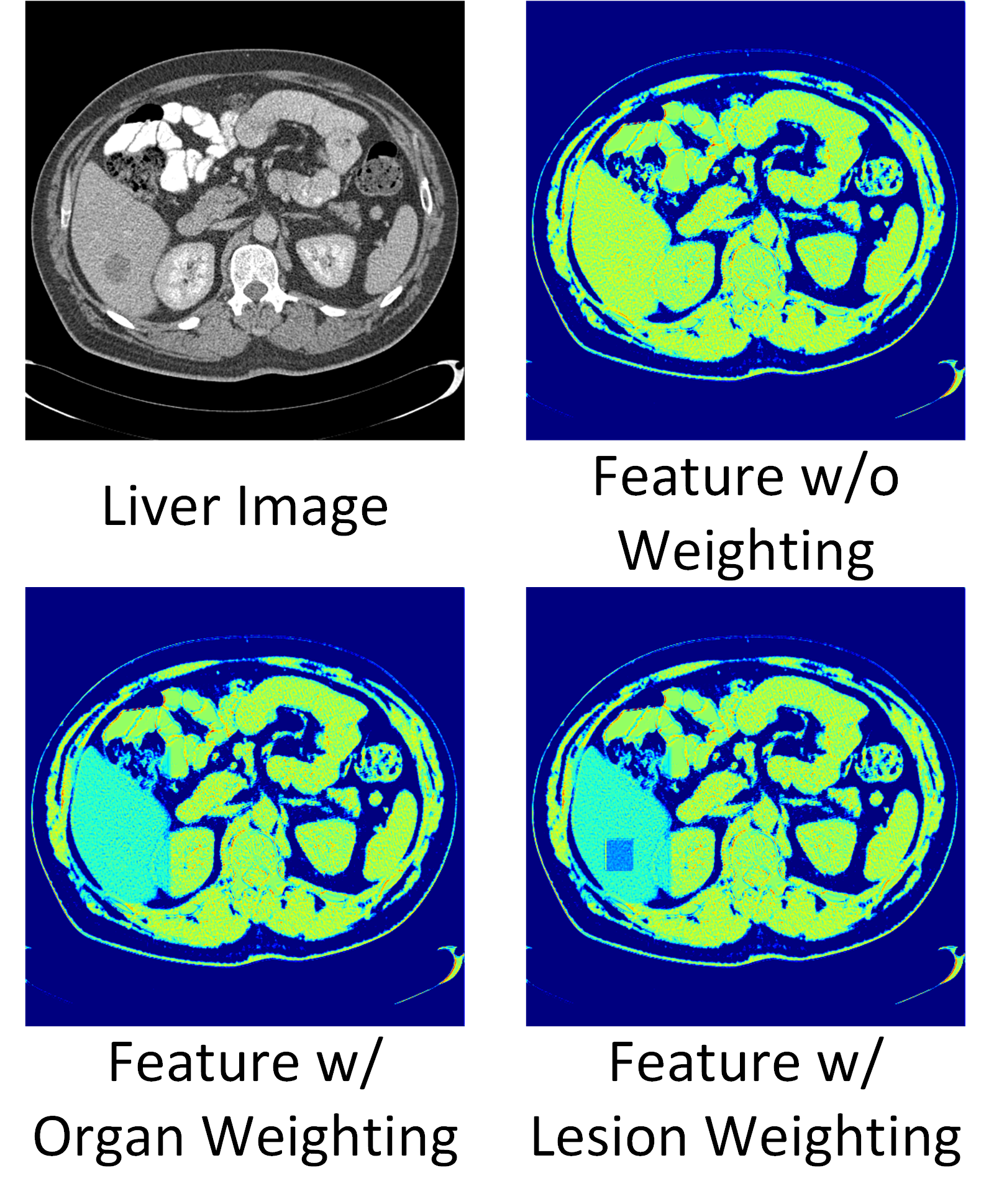}
	\caption{One visualized example to illustrate an O2L attention guide network to focus on organ and then lesion. We visualize a liver image, its visualized deep feature before weighting, after weighting by organ attention and lesion attention.}
	\label{Fig4}
\end{figure}

Cutting-edge Fully convolutional neural networks (FCN) were developed to achieve promising performances in different areas in medical image segmentation like brain parcellation \cite{ref1,ref17}, organ segmentation and lesion segmentation \cite{ref2,ref3,ref4,ref5,ref18,ref19}. A U-shape convolutional neural network called U-Net \cite{ref14,ref15} consists of a encoder and decoder with skip connection to help information flow in different depths. Various of modification of U-Net architecture were also proposed. Sun et.al. proposed a spatially weighted 3D U-Net for brain segmentation \cite{ref1}. A hybrid dense-UNet was used for liver segmentation in \cite{ref3}. In work of Seo et.al. \cite{ref2}, a object-dependence was modeled into a U-Net to enhance high level deep features. Bayesian learning was incorporated in a U-Net in \cite{ref20} for muscle segmentation. Reinforcement learning like Q learning was used in \cite{ref19} for pancreas segmentation. Besides a U-Net backbone, other deep architectures were also developed. A VNet was proposed for organ segmentation in \cite{ref4}. Deep supervision was used to enhance shallow features in \cite{ref5}. The above mentioned FCN model requires large amount of full annotated data pairs, limiting its wider application.

\subsection{Semi-supervised Image Segmentation}
In order to address the drawbacks of full supervision learning, semi-supervised learning leveraged massive unlabeled or weakly labeled datasets to compensate meager dense labeled data. As a form of weaker annotation compared with segmentation labels, bounding boxes provide rich information on object size and location yet little information on object shape. However, in practical labeling scenarios, labeling bounding box demands less efforts and thus leads to more efficient annotations. In the previous method called BoxSup \cite{ref21}, a iteration between generating region proposals automatically and training deep networks to gradually uncover underlying true segmentation masks. Papandreou et.al. \cite{ref22} proposed a weakly- and semi-supervised segmentation under an expectation-maximization framework. Similar to \cite{ref21}, an iterative refinements of segmentation mask based on GrabCut \cite{ref23} and handcrafted rules was proposed in \cite{ref24}. A box-driven class-wise masking model was incorporated into a semi-supervised learning framework combined with a filling rate guided adaptive loss to suppress meaningless and wrong annotations. With a self-correction network, weak set only containing bounding-box supervision promotes segmentation accuracy of a major segmentor in a recent work \cite{ref13}.

For medical image segmentation with mixed segmentation and bounding-box annotations, MS-Net \cite{ref8} was proposed for such mix supervision in FRRN network \cite{ref32} depending on its inputting supervision in a multi-task manner. In the work called MSDN \cite{ref9}, multiple backbones were applied for each kind of supervision for decoupling. However, both two methods utilized bounding-box supervision only by its localization and size information, and the shape information of weakly annotated data was overlooked. Besides, MS-Net and MSDN only performed two-classes segmentation without accounting for the condition of multiple-class segmentation like segmentation of background, organ and lesion.
A teacher-student model specifically designed for medical images is capable of generating high-quality pseudo labels rich in size, localization and shape information, thus boosting the multi-class segmentation accuracy close to a full supervision regime.

\section{Methods}
We start by a brief formulation of the semi-supervised learning of mixed supervision problem. Then we present the overall training scheme of a paired teacher-student model. We describe the teacher model producing high quality pseudo labels trained on limited number of dense label data. Then a student model learns from both manually segmentation labels and pseudo segmentation annotations by virtue of a localization branch.

\subsection{Problem Formulation}
\label{sec:PF}
\begin{figure*}[t]
	\centering
	\includegraphics[width = 0.9\textwidth]{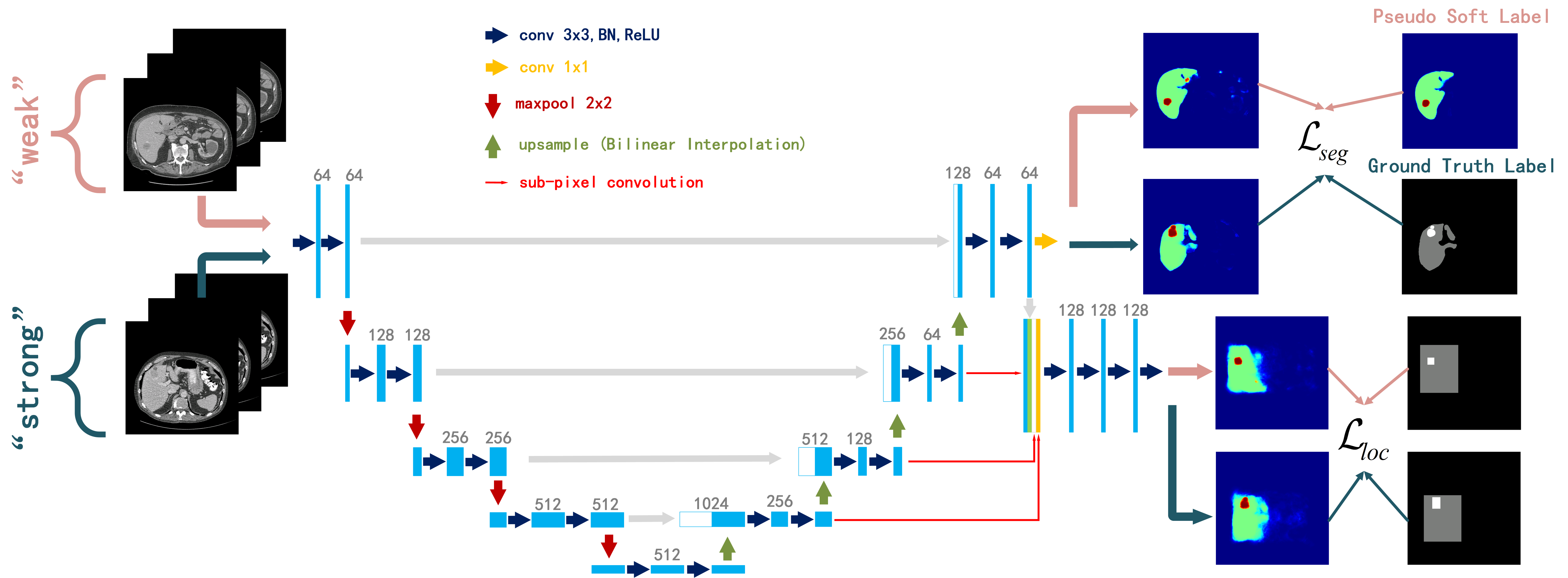}
	\caption{The deep network architecture of a student segmentor with localization branch by feature aggregation. The mixed datasets of augmented weak dataset ${\widetilde {\mathcal D}^w}$ and strong dataset ${\mathcal D}^s$ is used. Consecutive three slices are concatenated as network input.}
	\label{Fig6}
\end{figure*}
Suppose we have a training dataset with mixed supervision, we divide the dataset into two subdataset: one dataset called strong dataset ${\mathcal{D}^s} = \left\{ {x_i^s,y_i^s,b_i^s} \right\}_{i = 1}^N$ and another called weak dataset ${\mathcal{D}^w} = \left\{ {x_i^w,b_i^w} \right\}_{i = 1}^M$. There are total $N$ data triplets in a strong dataset $\mathcal{D}^s$ consisting of image $x^s$, corresponding segmentation annotation $y^s$ and bounding box $b^s$ encasing dense label $y^s$. In comparison, a weak dataset $\mathcal{D}^w$ contains $M$ data pairs only made up of image $x^w$ and bounding box $b^w$ of objects. Some examples are shown in Figure. \ref{Fig1}. Our purpose is utilizing the less informative mixed supervision to the most compared with a full-supervision one ${D^f} = \left\{ {x_i,y_i,b_i} \right\}_{i = 1}^{N + M}$. Note that for strong dataset, dense labels $y^s$ can be easily transformed into bounding box labels $b^s$ by finding circumscribed rectangles. For mutli-class segmentation of a medical slice, a bounding-box annotation $b$ consists of two types of bounding boxes: organ box $b_{og}$ and lesion box $b_{le}$. In most cases a lesion bounding box $b_{le}$ is a sub-region of organ bounding box $b_{og}$, which is shown in Figure. \ref{Fig1}. In the teacher and student network, we use hierarchical organ and lesion bounding boxes in different forms.

\subsection{Training Scheme}
We plot the whole training scheme in Figure. \ref{Fig2}. We first train a teacher annotator using a strong dataset $\mathcal{D}^s$, which is denoted in blue lines. Aided by a trained teacher, we augment a weak dataset $\mathcal{D}^w$ to an enriched ${\widetilde {\mathcal D}^w}$ containing soft pseudo dense labels. This forward pass is presented in green dashed lines. An augmented weak dataset ${\widetilde {\mathcal D}^w}$ in light red lines and strong dataset $\mathcal{D}^s$ in dark red lines are merged into a combination. The data combination teaches a student segmentor. Note bounding boxes $b^s$ and $b^w$ in ${\widetilde {\mathcal D}^w}$ and $\mathcal{D}^s$ are also utilized in the proposed localization branch in the student segmentor.

\subsection{Teacher Annotator}
\begin{figure*}[t]
	\centering
	\includegraphics[width = 0.9\textwidth]{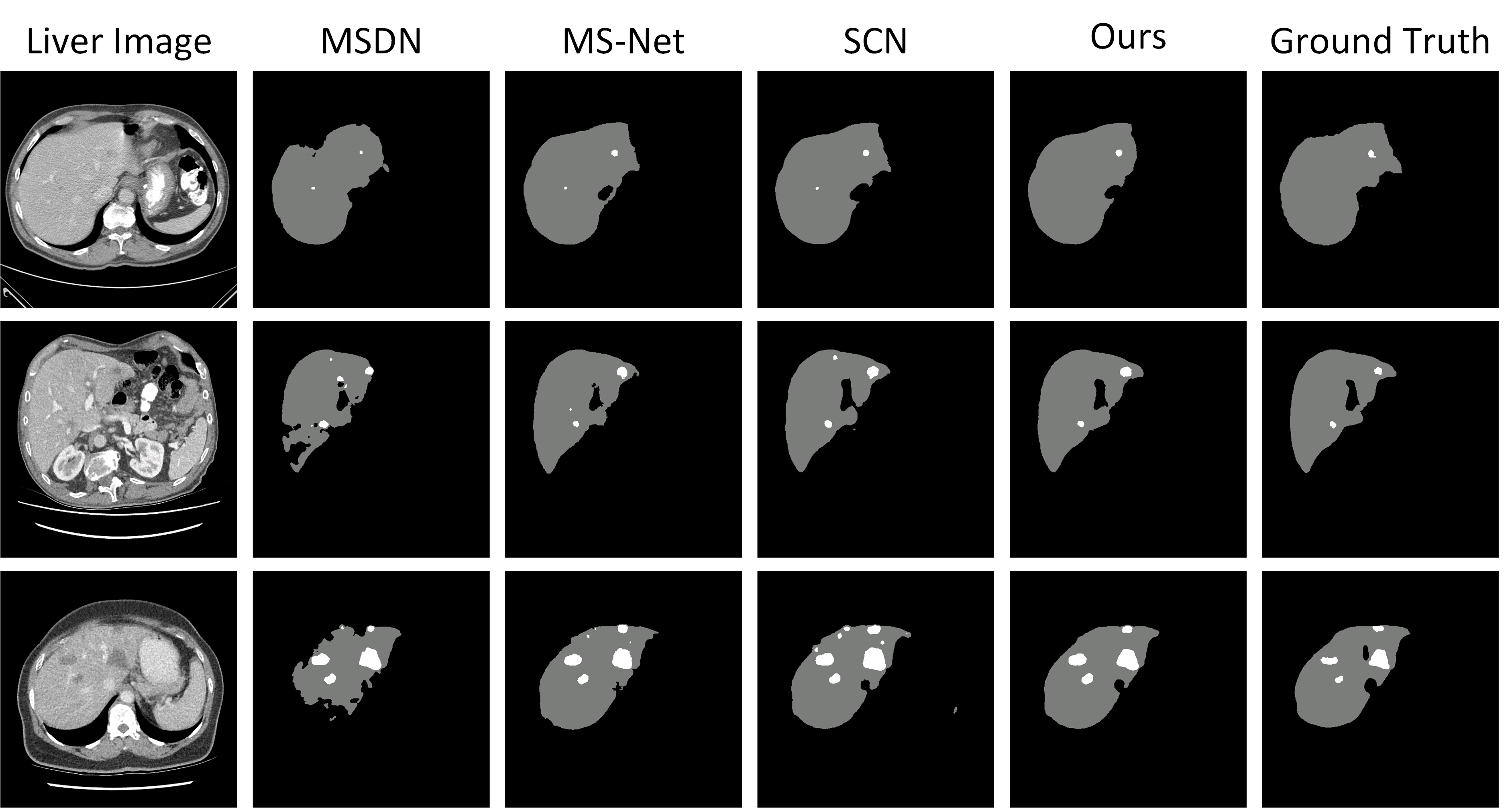}
	\caption{The segmentation results of the compared state-of-the-art semi-supervised learning methods including MSDN \cite{ref9}, MS-Net \cite{ref8} and SCN \cite{ref13}.}
	\label{Fig7}
\end{figure*}

Following the order in the training scheme, we first detail how we have a teacher annotator. Driven by the purpose of refining rough bounding-box supervision, a teacher annotator is expected to exploit shape information in limited number of strong annotated dataset $\mathcal{D}^s$ to prepare shape-aware pseudo segmentation labels on weak dataset for a student segmentor.

We built the teacher annotator based on a deep convolutional neural network. We show the architecture of a teacher annotator in Figure \ref{Fig3}. For the backbone, we adopt a U-Net \cite{ref14} for its effectiveness and efficiency as demonstrated in previous study \cite{ref29}. Also a standard U-Net architecture is amenable to add ad-hoc designs. Skip connections are used to concatenate shallow-layer features with deep-layer ones, which helps better fuse localization information in low-level features into deep layers rich in semantics \cite{ref26}. To better utilize spatial-temporary information, we concatenate three consecutive slices as a multi-channel input.

In addition to a vanilla U-Net, we observe a bounding box $b^s$ indicates where the object lies accurately.
Especially for tiny lesions which may only make up for a small portion in an unhealthy liver, accurate localization is crucial in detecting them. Therefore, to enhance shallow layers in a teacher net via such localization information, we proposed a hierarchical and progressive weighting strategy in a organ-to-lesion (O2L) manner. Motivated by the physiological fact that lesions in most cases exist within an organ, we assume that gradually focusing on an organ within an image, i.e. a liver within an abdomen image, and then its interior lesion brings benefits to joint segmentation of the organ and lesion.

In an O2L attention module as shown in Figure. \ref{Fig3}, features in each scale from the shortcut in an encoder is first recalibrated by weighting scores from an organ attention. Given a groups of features $f_{in}^j$ at the $j$th scale in an encoder and a binarized bounding box presenting organ $b_{og}^j$, we guide the network to focus on the organ by an organ attention
\begin{equation}
\label{eq1}
f_{og}^j = f_{in}^j + \sigma \left( {con{v_{og}^j}\left( {b_{og}^j} \right)} \right) \odot f_{in}^j,
\end{equation}
where $conv_{og}^j$ denotes convolutional layers at $j$th scale followed by batch normalization and $\sigma$ is a sigmoid activation function. A feature modulation is performed by a point-wise multiplication between input features and weighting scores. Note here liver and lesion box regions are merged together to form an organ region bounding box $b_{og}$ where entries in the union set are ones and zeroes elsewhere.

Motivated by the observation that lesions lie inside an organ, we further reweight the weighted features of an organ attention by a lesion attention. With a binarized bounding box presenting lesion $b_{le}^j$ and convolutional layers $conv_{out}^j$, a similar residual lesion attention is proposed to get output double-weighted features formulated as
\begin{equation}
\label{eq2}
f_{out}^j = f_{og}^j + \sigma \left( {con{v_{out}^j}\left( {b_{le}^j} \right)} \right) \odot f_{og}^j,
\end{equation}
where the lesion bounding box $b_{le}$ is a binary matrix in which the entries are ones on lesion box regions and zeros elsewhere.


Due to the various size of organs and lesions, we apply the hierarchical double weighting in each scale of the U-Net architecture. Here we adopt the residual attention architecture \cite{ref31} where shortcut is introduced to maintain a identical mapping while the sigmoid-activated attention select the most task-relevant information. In the work \cite{ref31}, the residual attention is shown to help optimization. In our network, there are 4 scales meaning $j$ can be $1,2,3,4$. We apply bilinear down-sampling on the binarized bounding boxes $b_{og}$ and $b_{le}$ before convolutions in scales smaller than image size. We show a group of examples of visualized feature maps in Figure. \ref{Fig4}, it is observed that hierarchical bounding boxes progressively target the network on organ and lesion. In later ablation study, we validate the effectiveness of the double-weighting residual attention compared with vanilla attention scheme using multi-channel inputs formed by concatenating organ bounding box $b_{og}$ and lesion bounding box $b_{le}$, which fails to utilize the hierarchy in geometrical pattern of organ and lesion.

A standard cross-entropy loss $\mathcal{L}_{t}$ function for optimizing the parameters of a teacher $\theta^{te}$ is used to train teacher annotator
\begin{equation}
\label{eq3}
\mathop {\arg \min }\limits_{\theta{^{te}}}  \sum\limits_{i = 1}^N { - \log p\left( {y_i^s|x_i^s,b_i^s;\theta{^{te}} } \right)}
\end{equation}

Note the teacher annotator is trained with a strong dataset ${\mathcal{D}^s} = \left\{ {x_i^s,y_i^s,b_i^s} \right\}_{i = 1}^N$ containing images, dense labels and bounding-box labels. A trained teacher annotator is capable of generating massive pseudo labels $\widetilde y^w$ conditioned on weak datasets ${\mathcal D}^w = \left\{ {x_i^w, b_i^w} \right\}_{i = 1}^M$ as inputs. Inspired by knowledge distillation \cite{ref27}, we choose soft labels from the output of the softmax layer of the teacher annotator as pseudo labels. Such smoothed soft labels reflect how confident a teacher is when producing pseudo dense labels. Compared with raw bounding-box labels, pseudo labels have more shape information with certainty while maintaining size and position information.

\subsection{Student Segmentor}
\begin{table*}[]
\caption{The comparison in objective indexes among MSDN, MS-Net, SCN and our proposed model as well as the full-supervision learning model. The Dice coefficients are given in percentile.}
\center
\begin{tabular}{|l|c|c|c|c|c|}
\hline
\multirow{2}{*}{Model} & \multicolumn{2}{c|}{Liver}        & \multicolumn{2}{c|}{Lesion}       & Tumor Burden \\ \cline{2-6}
                       & Dice global \% & Dice per case \% & Dice global \% & Dice per case \% & RMSE         \\ \hline
MSDN                   & 86.7           & 87.1             & 59.8           & 52.0             & 0.068          \\ \hline
MS-Net                 & 90.8           & 90.5             & 72.2           & 57.4             & 0.027          \\ \hline
SCN                    & 92.6           & 92.6             & 74.8           & 59.7             & 0.028          \\ \hline
Ours                   & \textbf{92.8}           & \textbf{93.1}             & \textbf{76.5}           & \textbf{60.0}             & \textbf{0.023}          \\ \hline
Full-Supervision       & {\textcolor{red}{93.9}}           & {\textcolor{red}{94.4}}             & {\textcolor{red}{79.2}}           & {\textcolor{red}{63.7}}             & {\textcolor{red}{0.020}}          \\ \hline
\end{tabular}
\label{Tab1}
\end{table*}

Provided a combination of soft pseudo-labeled dataset ${\widetilde {\mathcal D}^w} = \left\{ {x_i^w,\widetilde y_i^w,b_i^w} \right\}_{i = 1}^M$ and strong manually-annotated dataset ${\mathcal{D}^s} = \left\{ {x_i^s,y_i^s,b_i^s} \right\}_{i = 1}^N$, our goal is to train a student segmentor able to generalize well to unseen testing data. We use a similar U-Net architecture as in teacher annotator for backbone. In small object detection such as lesion segmentation given a whole medical slice, accurate localization is challenging yet important. In classic U-Net model. We observe bounding-box annotations in both ${\widetilde {\mathcal D}^w}$ and ${\mathcal{D}^s}$ offer precise localization information of organ and lesion.

To exploit these weak supervision information for accurate small object localization, we propose a feature aggregation branch in a student annotator to enhance its decoder with localization information. Different from the design of a teacher annotator, no bounding-box annotation is available during network inference in testing. Thus we use an independent branch to avoid requirements for bounding box in testing. During practical use where only accurate segmentation matters, the feature aggregation branch can be safely discarded without introducing more network parameters. A localization branch aggregates deep features in the decoder path. Pixel shuffle and subpixel convolution \cite{ref30} is used to aggregate features of different sizes in different scales. The multi-scale aggregation tackles with the variation in both organs and lesions. The network architecture of a student segmentor is shown in Figure. \ref{Fig6}.

For efficient inference, we regress organ and lesion bounding boxes in a standard segmentation fashion while maintaining localization supervision can be effectively back propagated to decoder layers. Different from the utilized two-stage form of bounding box supervision in a teacher annotator, we adjust the bounding box labeling for a 3-class (background, liver and lesion) semantic segmentation. We partition a bounding-box label of an image into different sub-regions corresponding to each class to generate a one-hot label.

The loss function $\mathcal{L}_s$ for optimizing the parameters of a student segmentor $\theta^{st}$ is made up of a cross-entropy loss $\mathcal{L}_{seg}$ for the main branch segmentation and another cross-entropy loss $\mathcal{L}_{loc}$ for regressing bounding box in a localization branch.
\begin{equation}
\begin{aligned}
\label{eq4}
{{\cal L}_{seg}} &= \sum\limits_{i = 1}^N { - \log p\left( {y_i^s|x_i^s;{\theta ^{st}}} \right)}  + \sum\limits_{i = 1}^M { - \log p\left( {\widetilde y_i^w|x_i^w;{\theta ^{st}}} \right)} \\
{{\cal L}_{loc}} &= \sum\limits_{i = 1}^N { - \log p\left( {\widehat{b}_i^{s}|x_i^x;{\theta ^{st}}} \right)}  + \sum\limits_{i = 1}^M { - \log p\left( {\widehat{b}_i^{w}|x_i^w;{\theta ^{st}}} \right)} \\
{{\cal L}_s} &= {{\cal L}_{seg}} + \alpha {{\cal L}_{loc}}
\end{aligned}
\end{equation}
Here $\widehat {b}_i^s$ or $\widehat {b}_i^w$ represent one-hot labels converted from bounding box annotations. The parameter $\alpha$ balances the loss of main-branch segmentation and localization, which is set to 1 empirically.


One benefit of the design of loss function $\mathcal{L}_{loc}$ is that we can adjust the localization branch to the characteristics of a dataset using class weighting. Taking the datasest for liver and lesion segmentation for example, a larger bounding box enclosing a whole organ tends to leave relatively large marginal errors around box boundaries. However, such marginal errors reduce for lesions. Therefore a larger weights on regression of lesion boxes and smaller ones on regression of organ boxes alleviates the resulting biases. Also small objects are more sensitive to accurate localization than larger objects. Empirically, we set the weight of the cross-entropy for lesion in localization loss to 1 and the one for liver to 0.1.

\section{Experiments}
\begin{figure*}[t]
	\centering
	\includegraphics[width = 1\textwidth]{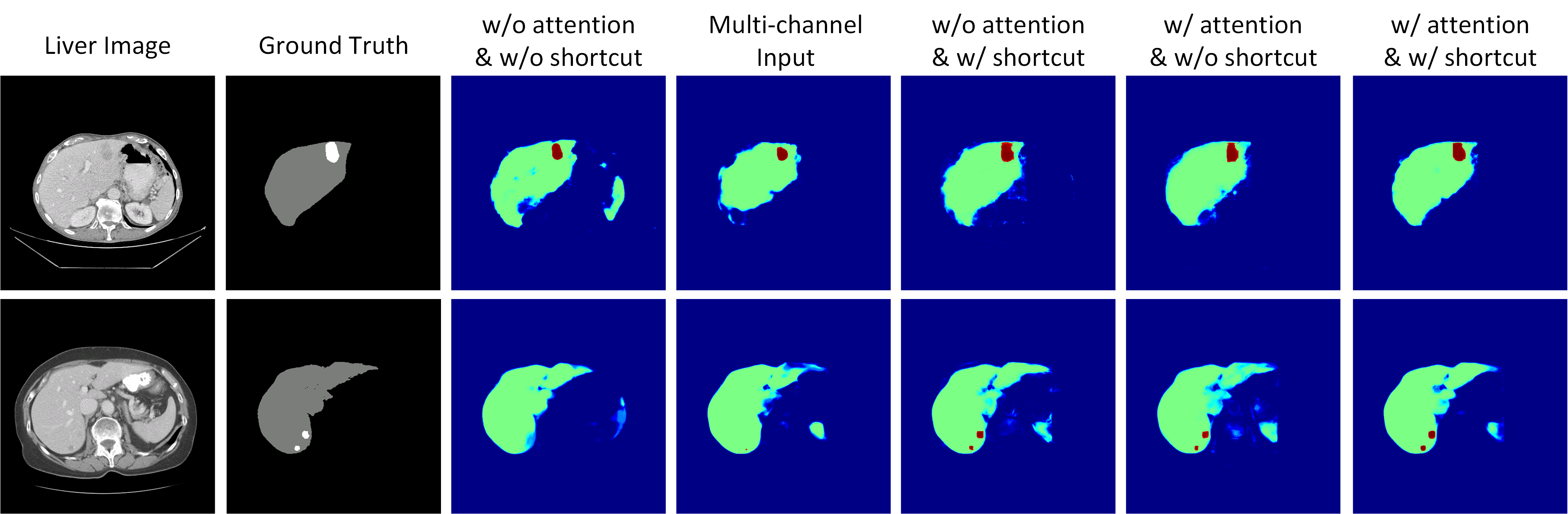}
	\caption{The visualized soft annotations produced by a teacher annotator in different settings in the ablation study of O2L attention weighting.}
	\label{TeaVis}
\end{figure*}

\subsection{Datasets}
We evaluate our model for organ and lesion segmentation on the publicized dataset of MICCAI 2017 LiTS Challenge\footnote{https://competitions.codalab.org/competitions/17094}. This dataset is aimed for liver and lesion segmentation. The LiTS dataset has 131 and 70 contrast-enhanced 3D abdominal CT scans for training and testing, respectively. Different scanners and protocols from six different clinical sites are used to construct this dataset. We preprocess the dataset by truncating HU intensities of all scans to the range of $[-200,250]$  to suppress irrelevant details. For data augmentation, we apply random scaling and flipping of the training slices between 0.8 and 1.2 following the work \cite{ref3}.

\subsection{Implementations}
We deploy our model on a RTX TITAN graphic card and use PyTorch as implementation platform. We use stochastic gradient descent (SGD) for optimization. For training of teacher annotator, the initial learning rate is set 1e-3 with momentum 0.9. The training takes total 1500 epoches. For training of student segmentor, the SGD is also used for optimization with initial learning rate 3e-4 and momentum 0.9. The training takes 4000 epoches. After the first 3000 epoches we decrease the learning rate by 0.95 in each epoch. The training ends with 4000 epoches. In each epoch we randomly crop a subvolume of the size $6\times320\times320$ from a CT scan. The subvolumes are transformed into groups of three consecutive slices for training. Mini-batchsize for training is set 2. Note in testing phase, we travel through all the consecutive three-slice groups in a testing scan with the localization branch removed from the student segmentor. Also in testing inference, a largest connected component labeling in a liver and hole filling in both liver and lesion are used for postprocess the segmentations, and the lesion segmentation is refined by removing lesions outside the liver regions.

\subsection{Evaluation Criterions}
Following the evaluation criterions used in the 2017 LiTS challenge, we use a Dice per case score and Dice global score to assess the whole liver and tumor segmentation performance. The Dice score is defined as
\begin{equation}
\label{eq4}
{\rm{DS = }}\frac{{{\rm{2}}\left| {{\rm{X}} \cap {\rm{Y}}} \right|}}{{\left| {\rm{X}} \right|{\rm{ + }}\left| {\rm{Y}} \right|}},
\end{equation}
where $\rm{X}$ and $\rm{Y}$ represent the prediction and ground truth. Dice per case score is an averaged Dice score calculated in a per-volume manner, and Dice global score denotes the Dice score evaluated on an unified dataset where all scans are combined together.

Besides the Dice Coefficients, we use root mean square error (RMSE) to measure the tumor burden of the liver. A higher Dice Coefficients and lower RMSE indicates a better segmentation.

\subsection{Comparison to the State of the Art}
\label{sota}
\begin{table}[]
\caption{The result of ablation study on the O2L attention weighting.}
\center
\begin{tabular}{|l|c|c|c|c|}
\hline
\multirow{2}{*}{Model}                                                   & \multicolumn{2}{c|}{Liver}                                                                                        & \multicolumn{2}{c|}{Lesion}                                                                                       \\ \cline{2-5}
                                                                         & \begin{tabular}[c]{@{}c@{}}Dice \\ global $\%$\end{tabular} & \begin{tabular}[c]{@{}c@{}}Dice \\ per case $\%$\end{tabular} & \begin{tabular}[c]{@{}c@{}}Dice \\ global $\%$\end{tabular} & \begin{tabular}[c]{@{}c@{}}Dice \\ per case $\%$\end{tabular} \\ \hline
\begin{tabular}[c]{@{}l@{}}w/o attention\\ \& w/o shortcut\end{tabular}  & 86.0                                                    & 85.3                                                      & 61.5                                                    & 42.9                                                      \\ \hline
\begin{tabular}[c]{@{}l@{}}multi-channel\\ input\end{tabular}            & 92.2                                                    & 92.6                                                     & 88.1                                                    & 83.9                                                     \\ \hline
\begin{tabular}[c]{@{}l@{}}w/o attention \\ \& w/ shortcut\end{tabular}  & 93.9                                                    & 93.3                                                     & 87.8                                                    & 85.2                                                     \\ \hline
\begin{tabular}[c]{@{}l@{}}w/ attention \\ \& w/o shortcut\end{tabular} & 94.0                                                    & 93.4                                                     & 88.8                                                    & 84.1                                                     \\ \hline
\begin{tabular}[c]{@{}l@{}}w/ attention \\ \& w/ shortcut\end{tabular}   & \textbf{95.4}                                                    & \textbf{95.2}                                                     & \textbf{88.9}                                                    & \textbf{85.5}                                                     \\ \hline
\end{tabular}
\label{Tab2}
\end{table}
Three other recently published works are compared with our model to demonstrate its state-of-the-art segmentation accuracy including MS-Net \cite{ref8}, MSDN \cite{ref9} and SCN \cite{ref13}. Among the 131 training scans in LiTS with publicized segmentation labels, $30\%$ percents are used as strong dataset with all images, bounding boxes and segmentations provided, and $60\%$ percents are used as weak dataset with only images and bounding boxes. The left $10\%$ percents are used for validation. The testing are performed on the 70 testing scans with segmentation ground truth held by challenge organizers. We show the quantitative segmentation results in Table. \ref{Tab1} using Dice global, Dice per case and RMSE as metrics. MS-Net outperforms MSDN. Because a teacher-student model was proposed to refine coarse bounding box annotations, the SCN achieve more accurate segmentation compared with MS-Net and MSDN. By virtue of the hierarchical O2L attention weighting in teacher annotator and localization branch in student segmentor, our model produces top accuracy among compared models and approximates the full-supervision learning approach.

We show three groups of segmentation results in Figure. \ref{Fig7}. As consistent with the quantitative results, our model produces better segmentation. It is notable that our model output less false positive predictions of lesions within a liver. This can be attributed to the help of the O2L attention and localization branch in detecting small lesions precisely.

\subsection{Ablation Study}
In this section, we validate the effective of the proposed hierarchical O2L attention in teacher annotator, the training of student network using pseudo labels and localization branch in student segmentor.

\subsubsection{Hierarchical O2L Attention in Teacher Annotator}
To evaluate the O2L attention in teacher annotator, we use $30\%$ of the 131 scans in the training dataset of LiTS for training and the left $70\%$ for testing. In our teacher annotator, we adopt a hierarchical double-weighting scheme with shortcut in each weighting. By removing attention weighting or shortcut connection, we compare the resulting models in Table \ref{Tab2}. We also compare our O2L attention mechanism with the attention strategy used in SCN \cite{ref13}, where two binarized bounding box annotations are concatenated as a multi-channel input for an attention module. In such a weighting strategy, the hierarchical geology of the medical images are underutilized. As shown in Table \ref{Tab2}, the model w/o attention and w/o shortcut is a lower bound in the comparison for no attention is leveraged. The attention model with multi-channel input is less accurate in predicting segmentations compared with other hierarchical models. Note the hierarchical attention model w/o shortcut achieves better segmentation on livers over the one without attention, demonstrating the utility of attention weighting. The model with both attention and shortcut connection achieves the best performance. In Figure. \ref{TeaVis}, we show two groups of the visualized pseudo soft segmentation produced by different teacher annotators. We observe the proposed hierarchical O2L attention helps improve segmentation. However, we still observe false prediction due to the limited number of training dataset.

\begin{figure}[t]
	\centering
	\includegraphics[width = 0.45\textwidth]{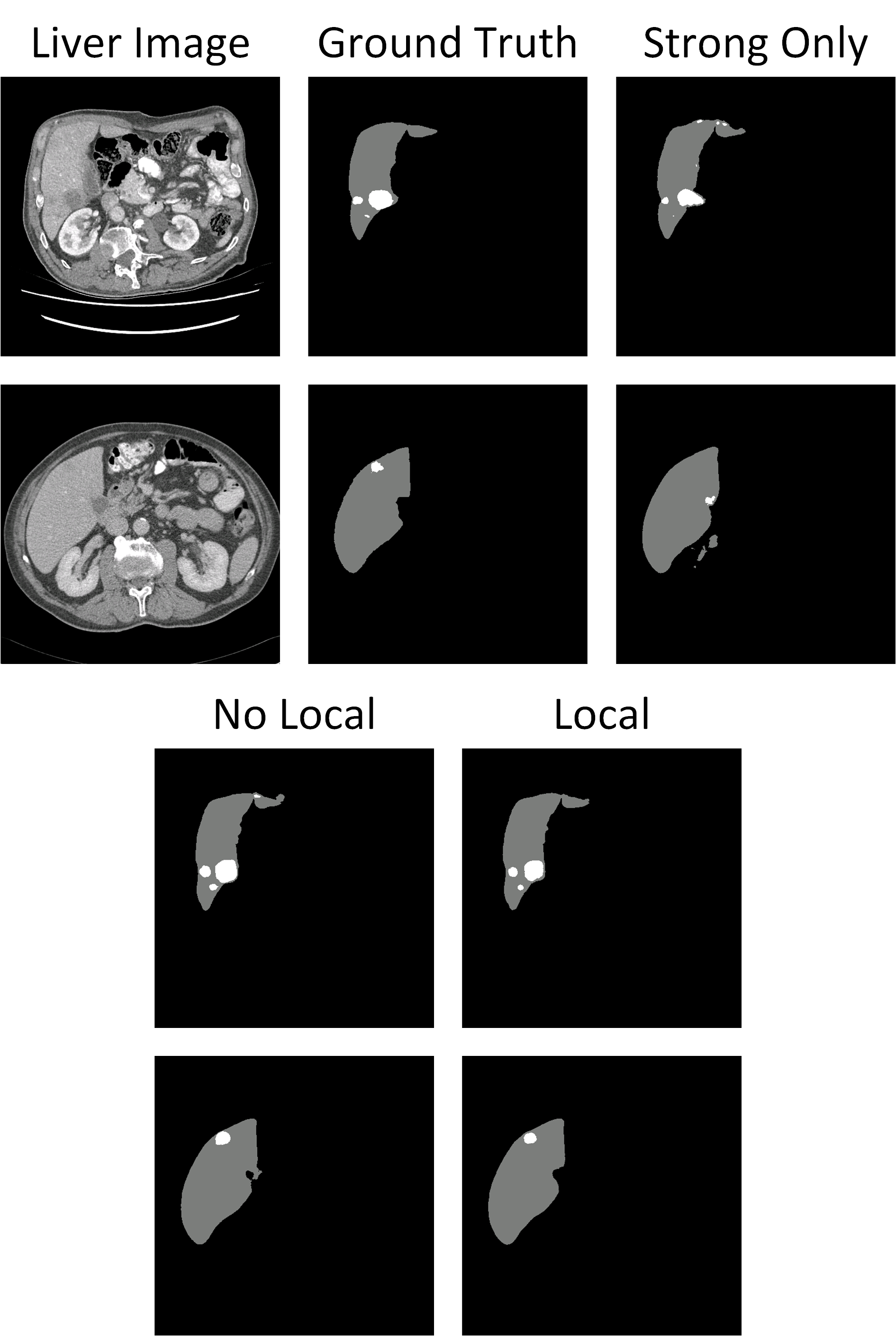}
	\caption{The visualized segmentations produced by a student segmentor in the ablation study of mixed supervision and localization branch.}
	\label{StuAbla}
\end{figure}

\subsubsection{Training Student in Mixed Supervision}
\label{TSMS}
\begin{figure}[t]
	\centering
	\includegraphics[width = 0.45\textwidth]{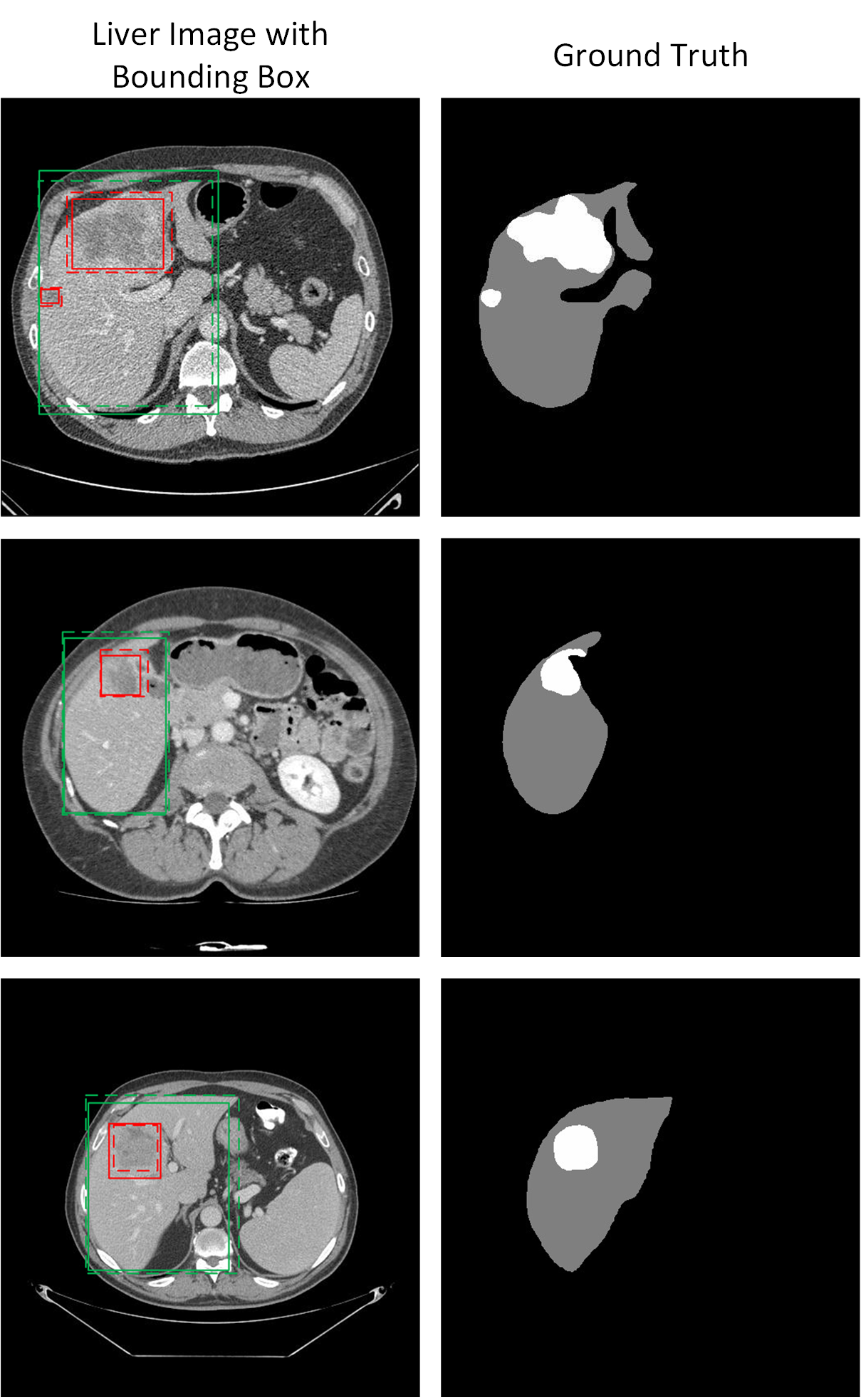}
	\caption{Three examples of the perturbed bounding box annotations. The dashed lines denote unperturbed golden standard boxes and the solid lines denote perturbed ones.}
	\label{PBB}
\end{figure}

\begin{figure}[t]
	\centering
	\includegraphics[width = 0.45\textwidth]{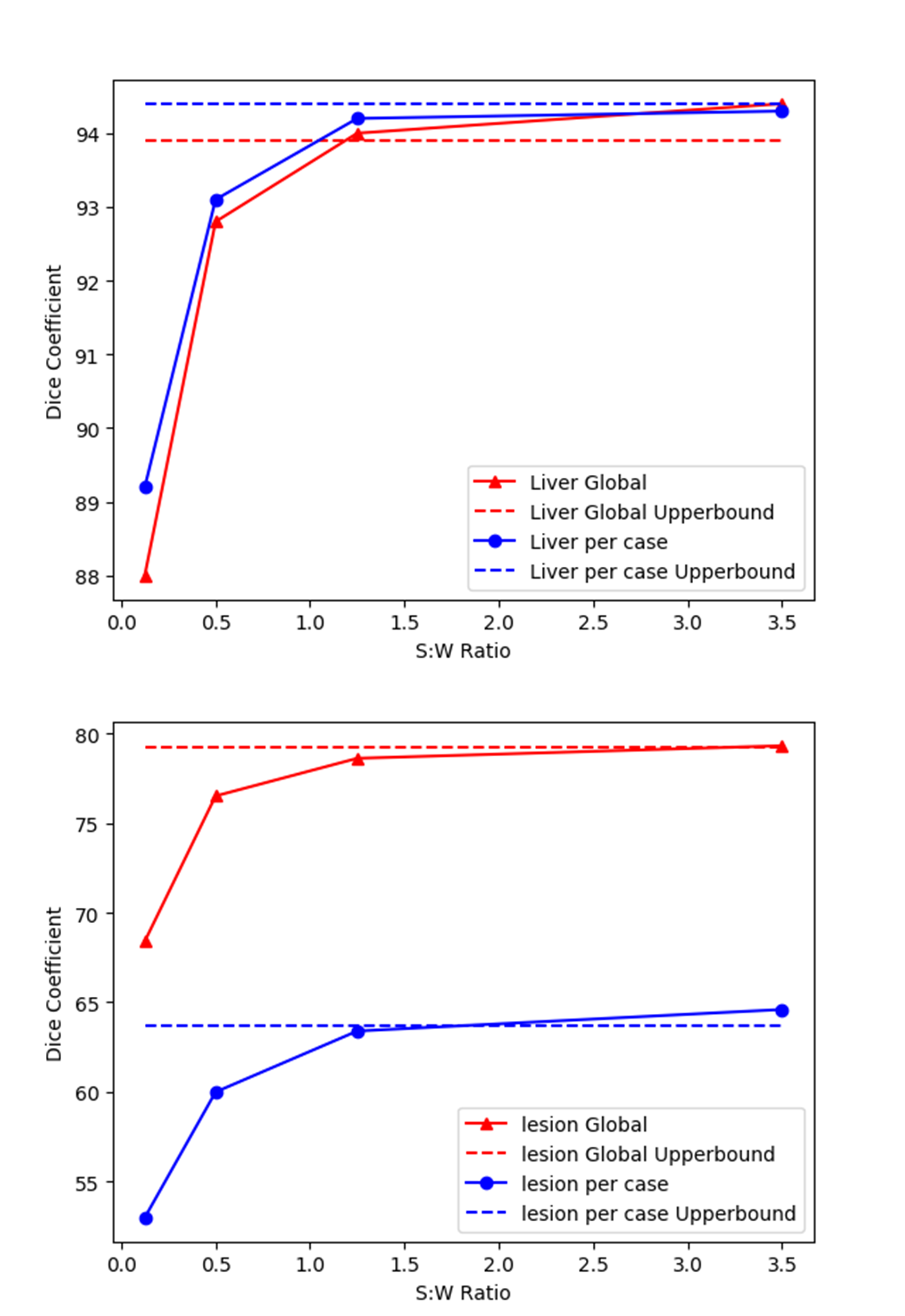}
	\caption{The performance curves of the Dice global and Dice per case for different settings of the ratios between the number of data in weak dataset and strong dataset.}
	\label{Ratio_Curve}
\end{figure}
We next evaluate how the student segmentor benefits from the pseudo labeling and compared with following two training strategy on the student segmentor without a localization branch: training on strong dataset only and training on mixed supervision. On a $30\%:60\%$ strong and weak dataset ratio, we evaluate the segmentation on a remaining $10\%$ scans. We find the student segmentor trained with strong dataset made up of only $30\%$ of the total amount of data achieves a Dice global $85.8\%$ and Dice per case $88.0\%$ on liver and a Dice global $61.8\%$ and Dice per case $43.9\%$ on lesion. In comparison, the student segmentor trained with the mixed strong dataset and pseudo-labeled weak dataset achieves a Dice global $92.6\%$ and Dice per case $93.2\%$ on liver and a Dice global $74.6\%$ and Dice per case $58.2\%$ on lesion. We observe a significant improvement on segmentation by virtue of the additional information from pseudo segmentation labels.

In a student segmentor, localization branch is a key component to guide its decoder with accurate localization information in bounding boxes labels. To evaluate the localization branch, we follow the same experiment setting in Section. \ref{TSMS} and add the localization branch in the student segmentor. We observe a Dice global $92.8\%$ and Dice per case $93.1\%$ on liver and a Dice global $76.5\%$ and Dice per case $60.0\%$. We observe the improvement on segmentation of lesion by a steady $2\%$, which is larger than the one on segmentation of liver, demonstrating our claim that accurate localization on small objects by bounding box supervision are crucial in their segmentation. We show some compared segmentation results in Figure. \ref{StuAbla}.

\subsection{Discussions on Quality of Bounding Box}
In our proposed semi- and weakly-supervised model, we transform the segmentation labels in partial dense-annotated data pairs into corresponding loose-labeled bounding-box annotations by finding their outermost points and drawing circumscribed rectangles as bounding boxes. Such less restrictive supervision is still precise enough to overlook the imperfect labeling often occurring in practical manual data annotation. In real scenarios, radiologists or labelers may annotate organs or lesions with physiological and medical prior knowledge in marking these bounding boxes while still introduce deviations or even errors for fast labeling. To test the robustness of our model to such efficient yet imperfect bounding box labeling, we randomly perturb a bounding box. Specifically, for bounding boxes of livers, we random scale a bounding box 0.95 to 1.1 times of the size, and then each of the four vertexes of the bounding box is randomly moved less than the $1/20$ of the corresponding sides. For bounding boxes of lesions within livers, we scale them 0.9 to 1.2 times of the size, and the vertexes are shifted less than the $1/10$ of the sides, due to their relative small size. We show some examples to illustrate such perturbations in Figure. \ref{PBB}.

In this simulation, similar to the Section. \ref{sota}, we use $30\%$ of the 171 scans as strong dataset, $60\%$ as weak dataset and the remaining $10\%$ for validation. The testing is conducted on the 70 testing scans in LiTS. On liver segmentation, our model achieves $92.5\%$ Dice global and $92.2\%$ Dice per case under the such perturbations. For lesion segmentation, the $76.5\%$ Dice global and $59.6\%$ Dice per case are achieved. Compared with the results we report in Section. \ref{sota}, we found the segmentation on lesions are almost unaffected by such perturbations and tolerant to the imprecision in bounding-box annotations, which may be attributed to the small size of lesions. The bias in bounding boxes decreases the accuracy in liver segmentation marginally. This experiments demonstrate the robustness of the proposed model and its potential utility in practical clinical scenarios.

\begin{figure*}[t]
	\centering
	\includegraphics[width = 0.65\textwidth]{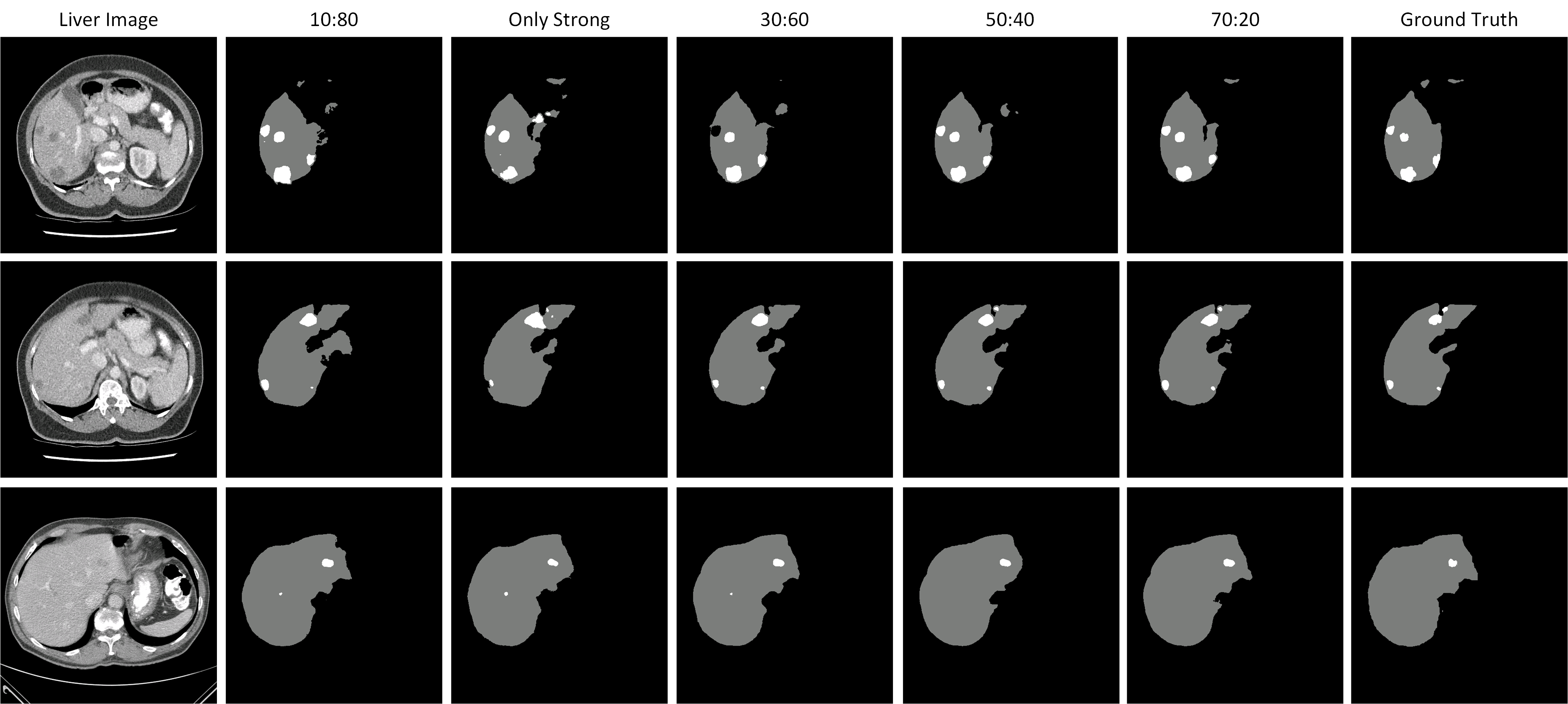}
	\caption{The segmentation results under different raito between the size of strong and weak dataset.}
	\label{Ratio_Vis}
\end{figure*}

\subsection{Discussions on the Ratio between Strong and Weak Dataset}
In this section, we validate our model under different ratios of the size between the strong and weak dataset. Reasonably, a higher ratio indicates that more data are labeled with pixel-wise dense supervision. In Figure. \ref{Ratio_Curve}, we show how the performance of our model varies with different ratios of strong and weak dataset. Among the 131 segmentation-label-publicized scans in the training dataset in the LiTS, $10\%$ of the scans are held out for evaluations and the remaining $90\%$ are split into a strong dataset and weak dataset with designed ratios such as $10:80$, $30:60$, $50:40$ and $70:20$. Note in Section. \ref{sota}, the ratio is adopted as $30:60$. We use Dice global and Dice per case as evaluation metrics. Note we also design a golden standard by setting the the ratio to $90:0$, meaning all the training data are dense annotated. In this upper bound, no teacher annotator is needed and a segmentor of the identical structure of a student segmentor is used except the localization branch. We observe a steady performance gain by gradually increasing the ratio. Surprisingly, our semi- and weakly-supervised learning model outperforms the golden-standard full-supervision model when the ratio exceeds $70:20$. This phenomenon demonstrates the high-qualities pseudo labels produced by a teacher annotator and the localization branch in a student segmentor contribute to the accurate segmentation. We show some segmentation results with different ratios in Figure. \ref{Ratio_Vis}.

\section{Conclusions}
In this paper, we proposed a semi- and weakly-supervised learning approach for medical image segmentation based on a teacher-student architecture to relax the demand for dense labeling. A teacher annotator produces high-quality pseudo segmentation labels by a hierarchical organ-to-lesion attention weighting. A student segmentor is trained by the combination of pseudo labels and manual labels and by a localization branch. Experimental results on the LiTS dataset for liver and lesion segmentation demonstrates the state-of-the-art performance of the proposed method, and our model achieves comparable performance compared with a full-supervision strategy. We also validate our model is robust to the perturbation in bounding box labeling, proving its potential application in real clinical practice.

\bibliographystyle{IEEEtran}
\bibliography{TSM}

\end{document}